\documentclass[sigconf]{acmart}
\usepackage{bbding}
\usepackage{threeparttable}
\usepackage{multicol}
\usepackage{multirow}

\newcommand{\hv}[0]{\ensuremath{\boldsymbol{h}} }

\newcommand{\xv}[0]{\ensuremath{\boldsymbol{x}} }

\newcommand{\thetav}[0]{\ensuremath{\boldsymbol{\theta}} }

\AtBeginDocument{%
  }

\copyrightyear{2025}
\acmYear{2025}
\setcopyright{acmlicensed}\acmConference[WWW Companion '25]{Companion
Proceedings of the ACM Web Conference 2025}{April 28-May 2, 2025}{Sydney,
NSW, Australia}
\acmBooktitle{Companion Proceedings of the ACM Web Conference 2025 (WWW
Companion '25), April 28-May 2, 2025, Sydney, NSW, Australia}
\acmISBN{979-8-4007-1331-6/2025/04}
\acmDOI{XX/XXXX}

\begin{document}

\title[SteamCast: Reliable Hail Nowcasting]{A Spatial-temporal Deep Probabilistic Diffusion Model for Reliable Hail Nowcasting with Radar Echo Extrapolation}



\author{Haonan Shi}
\authornote{Both authors contributed equally to this research.}
\affiliation{%
  \institution{School of Computer Science and Technology, Xidian University}
  \city{Chang'an Qu}
  \state{Xi'an Shi}
  \country{China}
}

\author{Long Tian}
\authornotemark[1]
\affiliation{%
  \institution{School of Computer Science and Technology, Xidian University}
  \city{Chang'an Qu}
  \state{Xi'an Shi}
  \country{China}
}


\author{Jie Tao}
\affiliation{%
  \institution{Hangzhou Institute of Technology, Xidian University}
  \city{Xiao Shan Qu}
  \state{Hang Zhou Shi}
  \country{China}
}

\author{Yufei Li}
\affiliation{%
  \institution{School of Computer Science and Technology, Xidian University}
  \city{Chang'an Qu}
  \state{Xi'an Shi}
  \country{China}
}

\author{Liming Wang \Envelope}
\affiliation{%
  \institution{School of Computer Science and Technology, Xidian University}
  \city{Chang'an Qu}
  \state{Xi'an Shi}
  \country{China}
}
\email{wanglm@mail.xidian.edu.cn}

\author{Xiyang Liu }
\affiliation{%
  \institution{School of Computer Science and Technology, Xidian University}
  \city{Chang'an Qu}
  \state{Xi'an Shi}
  \country{China}
}

\renewcommand{\shortauthors}{Haonan Shi et al.}

\begin{abstract}
  Hail nowcasting is a considerable contributor to meteorological disasters and there is a great need to mitigate its socioeconomic effects through precise forecast that has high resolution, long lead times and local details with large landscapes. Existing medium-range weather forecasting methods primarily rely on changes in upper air currents and cloud layers to predict precipitation events, such as heavy rainfall, which are unsuitable for hail nowcasting since it is mainly caused by low-altitude local strong convection associated with terrains. Additionally, radar captures the status of low cloud layers, such as water vapor, droplets, and ice crystals, providing rich signals suitable for hail nowcasting. To this end, we introduce a Spatial-Temporal gEnerAtive Model called SteamCast for hail nowcasting with radar echo extrapolation, it is a deep probabilistic diffusion model based on spatial-temporal representations including radar echoes as well as their position/time embeddings, which we trained on historical reanalysis archive from Yan'an Meteorological Bureau in China, where the crop yield like apple suffers greatly from hail damage. Considering the short-term nature of hail, SteamCast provides 30-minute nowcasts at 6-minute intervals for a single radar reflectivity variable, across 9 different vertical angles, on a latitude-longitude grid with approximately 1 km × 1 km resolution per pixel in Yan'an City, China.
  By successfully fusing the spatial-temporal features of radar echoes, SteamCast delivers competitive, and in some cases superior, results compared to other deep learning-based models such as PredRNN and VMRNN.
\end{abstract}




\maketitle

\section{Introduction}
Hail nowcasting is the task of predicting the occurrence, intensity, and location of hail events. This involves analyzing a variety of meteorological data, such as temperature profiles and radar reflectivity, to identify atmospheric conditions favorable for hail formation. The goal is to provide timely warnings to the public and relevant authorities so that they can take necessary precautions to mitigate the impacts of hail on agriculture, infrastructure, and public safety. Accurate hail forecasting is challenging due to the complex and rapidly changing nature of the atmospheric conditions that lead to hail, and it often necessitates advanced extrapolation techniques, encompassing numerical weather prediction systems \cite{ruzanski2011casa,pulkkinen2019pysteps} and data-driven machine learning algorithms \cite{ravuri2021skilful,zhang2023skilful}.

\begin{figure*}[t]
  \centering
  \includegraphics[width= 0.85\linewidth]{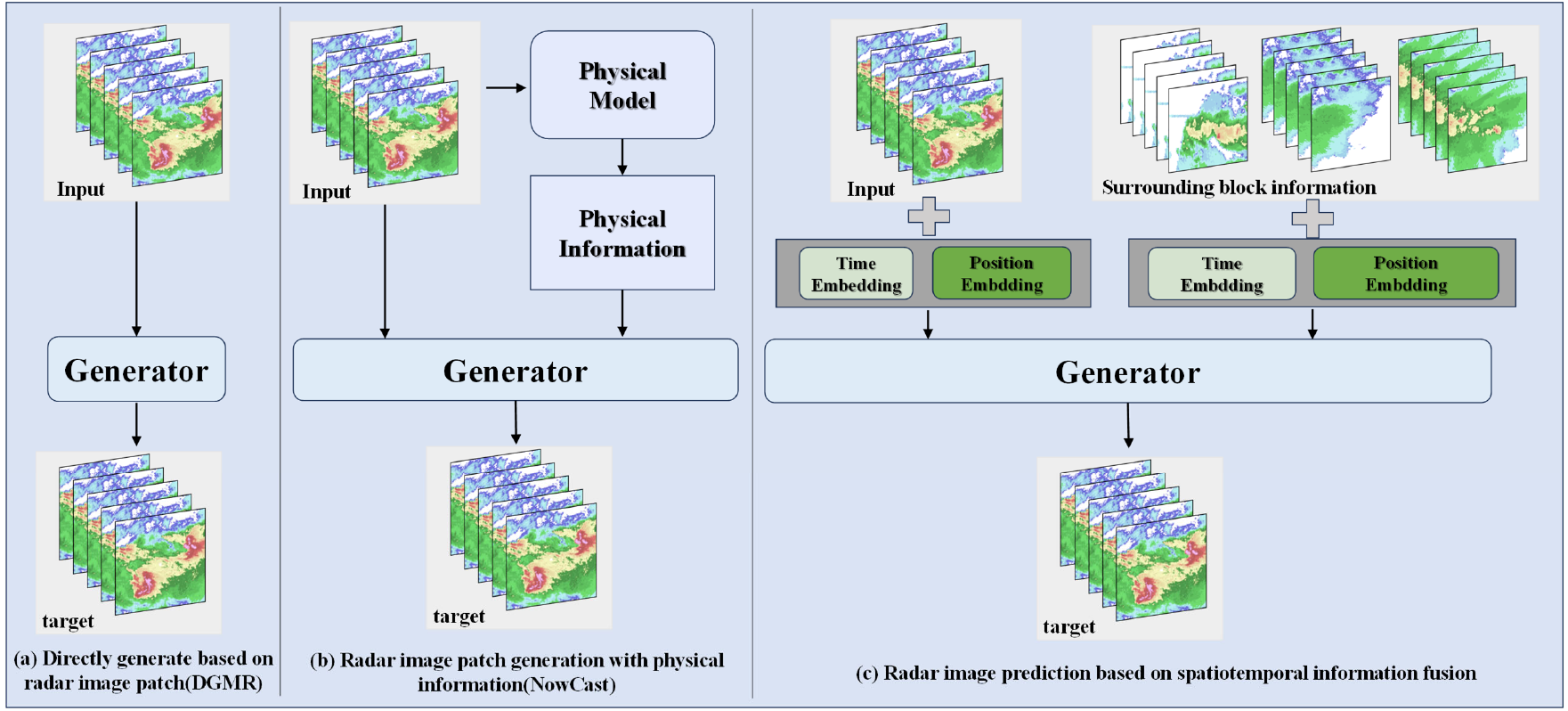}

  \caption{\small Comparison of our SteamCast in (c) with the DGM-based method in (a) and the physics-informed method in (b). Enhancing the capability of capturing spatial-temporal features is one of the most important aspects of reliable precipitation nowcasting. Methods (b) and (c) utilize a physical model
  and a well-designed spatial-temporal DGM to enhance the capability of capturing spatial-temporal features. 
  }
  \label{fig: moti}
\end{figure*}

Ensemble numerical weather prediction (NWP) systems often struggle to deliver accurate precipitation forecasts with a lead time of zero to two hours, primarily due to challenges in assimilating non-Gaussian data \cite{buehner2020non}.
As a result, alternative methods using radar data, such as PySTEPS \cite{pulkkinen2019pysteps} and DARTS \cite{ruzanski2011casa}, have been developed based on advection schemes, which are fundamentally rooted in the continuity equation.
However, current implementations of advection schemes fail to incorporate nonlinear evolution processes, leading to reduced accuracy in precipitation nowcasts, such as heavy rain reported in \cite{zhang2023skilful} and hailstorms analyzed in our study.

Deep-learning methods have been applied to precipitation nowcasting in recent years \cite{wang2022predrnn,espeholt2022deep,wang2023skillful,shi2015convolutional}. These methods aim to better capture traditionally challenging non-linear precipitation phenomena, such as convective initialization and extreme precipitation, by leveraging large datasets of radar echoes to train neural networks, instead of relying on physical assumptions. This end-to-end data-driven optimization effectively reduces inductive biases.
Consequently, the nowcast quality of these methods improves, as reflected in metrics like the per-grid-cell Critical Success Index (CSI). However, these methods encounter increasing uncertainty at longer lead times, resulting in blurrier precipitation fields, which are critical for accurate nowcasts of small-scale radar echo patterns. We speculate that poor spatial-temporal representations limit the models' operational utility, preventing them from simultaneously providing consistent predictions across multiple spatial and temporal scales. A large step forward in enhancing spatial and temporal representations has been the Deep Generative Model (DGM) of Radar called DGMR \cite{ravuri2021skilful} developed by DeepMind and the UK Met Office. DGMR is also the first work employing DGM for precipitation nowcasting. Long etal. \cite{zhang2023skilful} propose NowcastNet, a reliable nowcasting framework that aggregates spatial-temporal features by integrating DGM with physical knowledge of precipitation processes, including the conservation law of cloud transport \cite{ruzanski2011casa} and the rain rate distribution \cite{crane1990space}. 
In contrast to physically informed spatial-temporal aggregation, we propose an alternative approach by extracting spatial-temporal features of radar echoes using a well-designed Spatial-Temporal gEnerAtive Model called SteamCast.

We adopt a stable diffusion architecture \cite{rombach2022high} with slight yet critical modifications to represent effective and flexible spatial-temporal radar echoes.
We employ a lightweight condition encoder to extract spatial-temporal features from a target patch and its neighboring regions in eight directions for hail nowcasting. A stable diffusion U-Net layer then merges the randomly initialized patch with these features to generate 30-minute nowcasting results. Leveraging cross-attention, our model effectively captures radar echoes and adapts to forecasting tasks with flexible input ($N$) and output ($M$) lengths. Additionally, to enhance efficiency under limited computing resources, we process radar images in smaller sub-problems instead of full $H \times W$ resolution, seamlessly aggregating the results.
In summary, SteamCast introduces a stable diffusion-based architecture with an efficient spatial-temporal feature extraction mechanism, enabling flexible and high-resolution hail nowcasting.

\section{Related Work}

One of the earliest deep learning-based methods is the ConvLSTM Network for precipitation nowcasting \cite{shi2015convolutional}, in which the authors framed nowcasting as a spatial-temporal sequence-to-sequence problem and proposed a cascaded Convolutional LSTM for its solution. Similarly, PredRNN \cite{wang2022predrnn} introduced a spatial-temporal memory flow mechanism for complex pattern nowcasting. More recently, DGM-based methods have gained significant attention for addressing this challenge task \cite{zhang2023skilful,wang2023skillful}, as previously discussed. In this work, we focus on hail nowcasting, which represents a novel and significant task. Unlike heavy rainfall, hail formation is strongly associated with terrains, making precise modeling of spatial-temporal correlations crucial for hail nowcasting, a topic that has been rarely studied.

\begin{figure*}[t]
  \centering
  \includegraphics[width=0.98\linewidth]{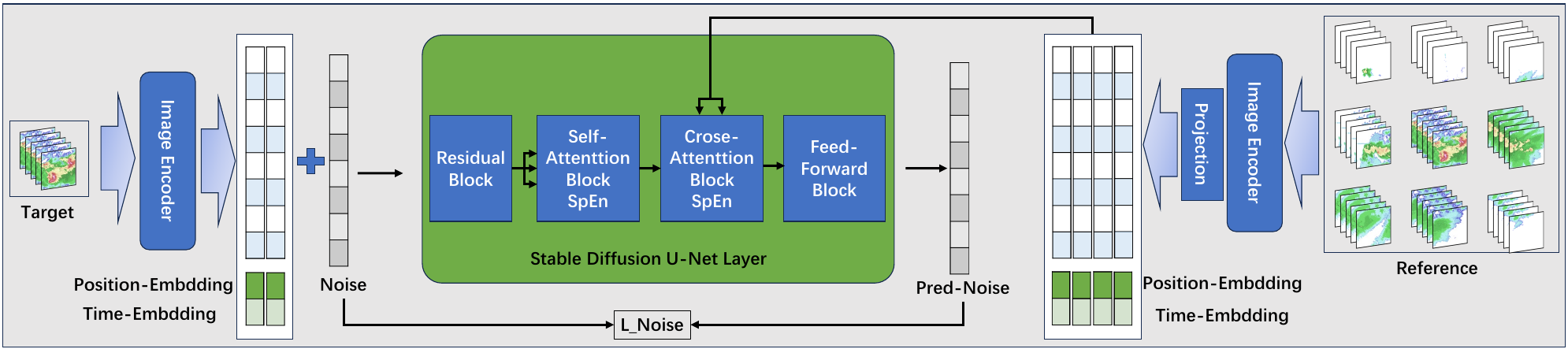}
  \caption{\small SteamCast architecture details. SteamCast adopts a SD architecture with slight yet critical modifications to effectively and flexibly represent spatial-temporal radar echoes. Left: The pre-trained encoder in SD captures features from the target patch of radar echoes, the target patch consists of $M$ nowcasted time steps. Middle: In U-Net, we apply self-attention within the target patch to encourage target-to-target consistency. We also apply a cross-attention block between the target patch and $T$ reference patches to enhance reference-to-target consistency. In each attention block, SpEn is employed for the key and query, enabling the attention map to capture relative positions and time steps. Right: The pre-trained encoder in SD and projection layer capture features from $T$ reference patches, each patch contains $N$ historical time steps.}
  \label{fig: steamcast}
\end{figure*}

\section{Methodology}
\subsection{Task Definition}
We consider reference patches of radar echoes denoted by $\{\xv_{i}^{r}\}_{i=1}^T$, where $\xv_i^r \in \mathbb{R}^{T \times H \times W \times N}$. Here $T$, $H$, $W$, and $N$ separately denote the number of reference patches, patch height, patch width, and historical time steps. Similarly, the target patch is denoted by $\xv_{j}^t \in \mathbb{R}^{1\times H \times W \times M}$ corresponding to the $j$-th reference patch. Here $j=\{1,...,T\}$ and $M$ is the number of nowcasted time steps. Our goal is to train a nowcasting model $\hat{\xv}_j^t=f_{\thetav}(\{\xv_{i}^{r}\}_{i=1}^T)$ such that $\|\xv_j^t - \hat{\xv}_j^t\| \leq \epsilon$, where $\epsilon$ is a small positive value. We randomly select a target patch from the reference patch set to train the model.


\subsection{SteamCast Architecture}

The architecture details of SteamCast, as shown in Fig. \ref{fig: steamcast}, are designed based on two key principles: 1) It leverages existing image-based diffusion models to inherit their powerful generalization capabilities. 2)  It encodes spatial-temporal information for each tokenized input patch, enabling the model to process an arbitrary number of patches and time steps for flexible nowcasting.

\noindent \textbf{Target Patch Generation 
:} SteamCast is designed by adopting the architecture of the Stable Diffusion v1.5 (SD) \cite{rombach2022high}. To adapt SD, which was originally designed for text-to-image generation, for hail nowcasting using radar extrapolation, several key modifications are implemented. First, we reimplement the self-attention block in the original SD to calculate interactions within the target patch across $M$ different nowcasted time steps,  thereby enhancing target-to-target consistency.
For the cross-attention block in the original SD, we modify it to calculate interactions from $T$ reference patches to the target patch, ensuring reference-to-target consistency. The features extracted by the SD encoder are $\hv_j^t \in \mathbb{R}^{b \times d}$, where $b=(h \times w) \times M$, with $h \times w$ is the total number of tokenized features.

\noindent \textbf{Reference Patches Condition:}
In hail nowcasting, it is crucial that the conditioning features accurately capture the texture details present in the reference patches, which have often been overlooked in previous studies. In SteamCast, we treat reference patches as sets of tokens and use an image encoder to extract such conditioning signals. This approach enables us to maintain flexibility in handling a variable number of patches and time steps. We choose to compress the reference patches to smaller resolution features with ConvNeXtv2-Tiny \cite{woo2023convnextv2}, a lightweight and highly efficient CNN architecture. The features of the reference patches after CNN are expressed as $\hv_i^r \in \mathbb{R}^{b^{\prime} \times d}$ for $i=1,...,T$, where $b^{\prime}= (h^{\prime} \times w^{\prime}) \times N$, with $h^{\prime} \times w^{\prime}$ the total number of tokenized features. 
The features after the self-/cross-attention block with SpEn can be written as $\hat{\hv}_j^t, \hat{\hv}_j^t \in \mathbb{R}^{b \times d}$. Details on SpEn are provided in Sec. \ref{spen}.

\noindent \textbf{Model Training and Inference:}
We use the original SD objective, which is to reconstruct the error, to fine-tune the parameters of SteamCast, which is initialized with pre-trained SD v1.5 \cite{rombach2022high} and ConvNeXtV2-Tiny. During inference, SteamCast takes a set of reference patches with $N$ historical time steps and a randomly initialized target patch with $M$ nowcasted time steps. Then it provides users with the corresponding nowcasting for the target patch of interest.


\subsection{Spatiotemporal Encoding (SpEn)}
\label{spen}

To efficiently encode spatial-temporal information into the reference and target patches in the U-Net layer of SD, we propose SpEn, inspired by advancements in the language domain. For illustration, consider the $i$-th reference patch at the $n$-th time step, denoted as $\hv^r_i(:,n)$. Suppose there are $4 \times 4=16$ reference patches and $5+5=10$ time steps, with $5$ for historical time steps and $5$ for nowcasting. Time and position indices can be separately represented by 4-D binary vectors ($\lceil log2K \rceil$). We employ the standard Sin/Cos embedding scheme used in Transformers \cite{vaswani2017attention} to encode each index (time or position), resulting in an encoded vector $\boldsymbol{\rho} \in \mathbb{R}^{1 \times 8}$. The final spatial-temporal embedding is obtained by repeating $\boldsymbol{\rho}$ to match the length of $\hv_i^r(:,n)$ along the last dimension. Hence, the tokenized feature after SpEn can be expressed as $\hat{\hv}_i^r(:,n) = \boldsymbol{\rho} \odot \hv_i^r(:,n)$, where $\odot$ represents element-wise multiplication.

\begin{figure}[t]
  \centering
  \includegraphics[width= 0.80\linewidth]{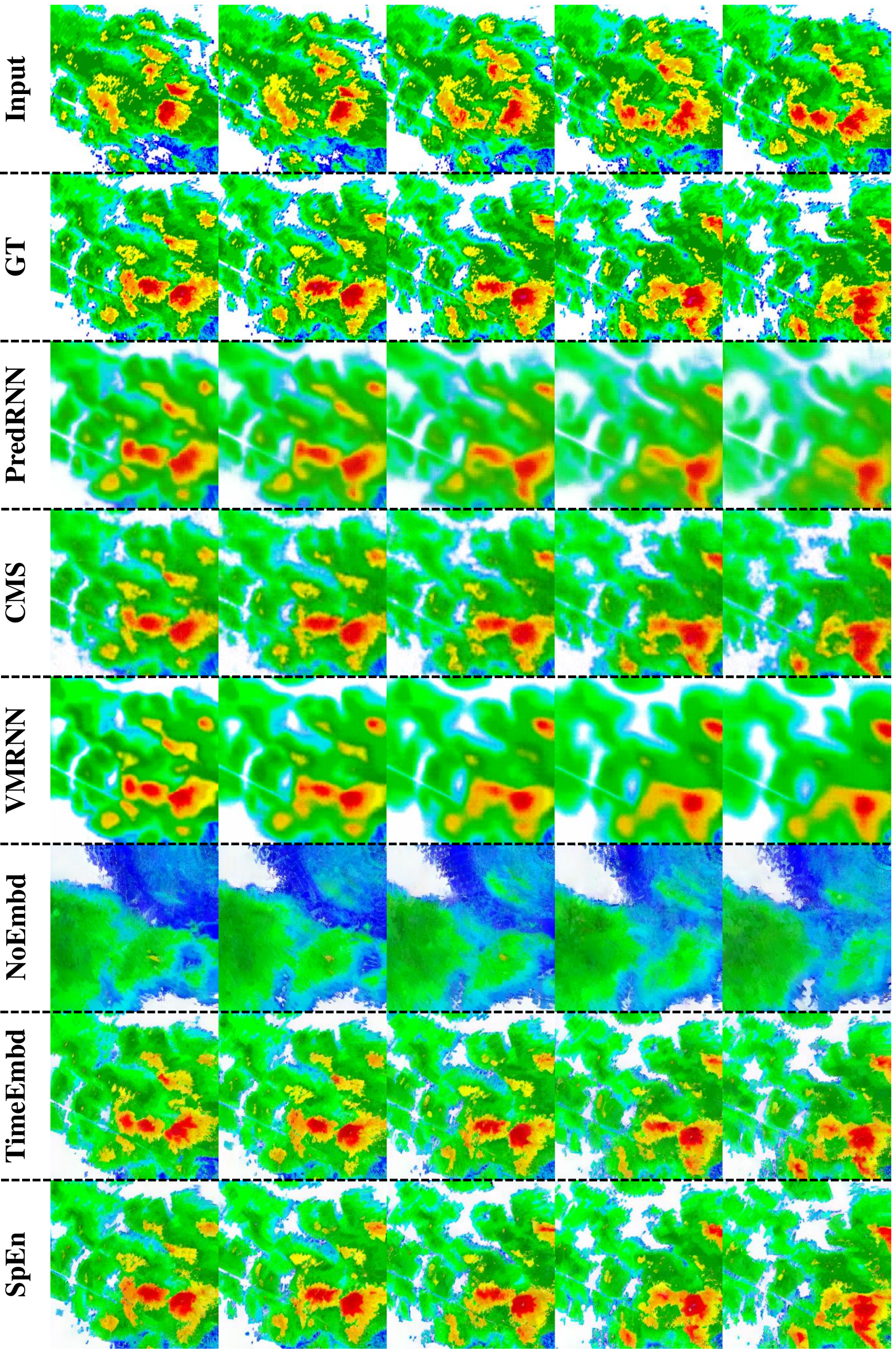}
  \caption{
  \small Visualization results of hail nowcasting. Input and GT denote target patches of 5 historical and 5 future time steps, respectively. PredRNN, CMS, and VMRNN represent nowcasting results from the respective competitors. SpEn represents the hail nowcast of SteamCast. NoEmbd and TimeEmbd represent hail nowcasts from our two variants, as discussed in Sec. \ref{sec:ablation} of the ablation studies.
  }
  \label{fig:vis}
\end{figure}

\section{Experiments}

\subsection{Experimental Setups}

\noindent \textbf{Dataset:} 
We use the 2-year historical reanalysis archive of radar reflectivity measurements with a resolution of 1024$\times$1024$\times$3 from Yan'an Meteorological Bureau in China to validate the performance of SteamCast. The interval between two adjacent measurements is 6 minutes. We collect a total of 10,000 radar reflectivity patches using a sliding window with size of $10 \times 256 \times 256 \times 3$ without overlapping across the entire historical dataset. Then, we randomly split the entire set of patches into 6,400, 1,600, and 2,000 patches for training, validation, and testing, respectively.

\noindent \textbf{Competing Methods:} 
We compare our SteamCast with some of the most recently proposed spatial-temporal prediction methods, including 1) PredRNN \cite{wang2022predrnn}: A spatial-temporal recurrent network. 2) CMS \cite{chai2022cms}: A memory-based convolutional network. 3) VMRNN \cite{tang2024vmrnn}: A memory-based recurrent network with variational autoencoding.

\noindent \textbf{Evaluation Protocols:} 
We use the following metrics to evaluate performance, including 1) MSE (Mean Squared Error): Pixel-wise difference between the ground truth and predicted target patches. 2) PSNR (Peak Signal-to-Noise Ratio): Reflects the quality of the predicted target patches, with higher values indicating better quality. 3) SSIM (Structural Similarity Index): Assesses the structural similarity between the ground truth and predicted target patches, considering luminance, contrast, and texture. 4) ETS (Equitable Threat Score): Measures the prediction accuracy for rare event predictions. 5) ACC (Accuracy): Measures the overall prediction accuracy.


\noindent \textbf{Implementation Details:} 
We use the encoder and decoder of SD v1.5 \cite{rombach2022high} to separately extract features of the target patch and generate the nowcasted target patch during training. We use ConvNeXtv2-Tiny \cite{woo2023convnextv2} to extract features of the set of reference patches. The U-Net layer is a modified version of SD v1.5. We use the pre-trained parameters of SD v1.5 and ConvNeXtv2-Tiny to initialize our SteamCast. Our model is trained with the following hyper-parameters: batch size 4, learning rate $1 \times 10^{-4}$, AdamW optimizer \cite{loshchilov2019decoupled}, training steps 100,000. Additionally, we use mixed precision training for acceleration. The experiments are conducted with two NVIDIA GTX 4090 GPUs.

\begin{table}[t]
\centering
  \caption{Nowcasting performance on various metrics. The best results are in bold.}
  \resizebox{0.45\textwidth}{!}{
  \begin{tabular}{c|c|c|c|c|c}\toprule
   \multicolumn{1}{c|}{\textbf{Model}} &\multicolumn{1}{c|}{\textbf{MSE} $\downarrow$} & \multicolumn{1}{c|}{\textbf{PSNR} $\uparrow$} & \multicolumn{1}{c|}{\textbf{SSIM} $\uparrow$} & \multicolumn{1}{c|}{\textbf{ETS} $\uparrow$} & \multicolumn{1}{c}{\textbf{ACC} $\uparrow$}\\  \midrule
   PredRNN (TPAMI'22) &0.05  &13.86 &0.62 &0.08 &0.96 \\ 
   CMS (ICME'22) &0.06  &13.09 &0.60 &0.07 &0.96 \\ 
   VMRNN (CVPR'24) &0.03  &17.02 &0.79 &0.10 &0.97 \\ 
   SteamCast (Ours) &\textbf{0.02}  &\textbf{23.15} &\textbf{0.81} &\textbf{0.18} &\textbf{0.99} \\ \bottomrule
  \end{tabular}}
  \label{table1}
\end{table}

\begin{table}[t]
\centering
  \caption{Ablation studies on spatial and temporal embeddings with various metrics. The best results are in bold.}
  \resizebox{0.45\textwidth}{!}{
  \begin{tabular}{c|c|c|c|c|c|c}\toprule
   \multicolumn{1}{c|}{\textbf{Position}} & \multicolumn{1}{c|}{\textbf{Time}} &\multicolumn{1}{c|}{\textbf{MSE} $\downarrow$} & \multicolumn{1}{c|}{\textbf{PSNR} $\uparrow$} & \multicolumn{1}{c|}{\textbf{SSIM} $\uparrow$} & \multicolumn{1}{c|}{\textbf{ETS} $\uparrow$} & \multicolumn{1}{c}{\textbf{ACC} $\uparrow$}\\  \midrule
   - &- &0.40  &10.00 &0.43 &0.01 &0.97 \\ 
  - &\checkmark &0.05  &\textbf{23.21} &0.74 &0.16 &0.98 \\ 
  \checkmark &\checkmark &\textbf{0.02}  &23.15 &\textbf{0.81} &\textbf{0.18} &\textbf{0.99} \\  \bottomrule
  \end{tabular}}
  \label{table2}
\end{table}
\subsection{Enumerated Results}

\noindent \textbf{Quantitative results:}
We compare SteamCast with recent state-of-the-art methods, and the results are presented in Table~\ref{table1}. All competing models are designed for spatial-temporal prediction, aiming to nowcast the target patch for the next 30 minutes based on historical observations. SteamCast consistently outperforms other methods across various metrics, demonstrating its effectiveness in hail nowcasting.
Given the rarity of hail events, ACC is not a reliable metric for evaluating predictive improvements. Instead, the Equitable Threat Score (ETS), which accounts for both true positives and random chance, provides a more meaningful assessment. SteamCast achieves a notable ETS increase from 0.10 to 0.18, demonstrating a substantial enhancement in hail detection capability.

\noindent \textbf{Qualitative results:}
We report visualized nowcasting results of a hail event in Fig. \ref{fig:vis}. As we can see, our model achieves promising performance compared with others not only in correct trends but also in richer details.

\subsection{Ablative Analysis}
\label{sec:ablation}

We emphasize the importance of properly aggregating spatial-temporal information of the target patch. To achieve this, we designed two additional variants of SteamCast: 1) NoEmbd: without time and position embeddings. 2) TimeEmbd: employing only time embedding. According to the results in Table \ref{table2}, we find that leveraging time and position embeddings contributes significantly to improved performance. Additionally, qualitative results in Fig. \ref{fig:vis} further support, in a more intuitive way, that SpEn is indispensable.

\section{Conclusion}

Our study demonstrates that SteamCast, a spatial-temporal deep probabilistic diffusion model, improves the accuracy and adaptability of hail nowcasting, with applications in precision agriculture, disaster prevention, and urban resilience. Future work will focus on integrating multi-source meteorological data to enhance long-term forecasting capabilities.

\begin{acks}
This work was supported by the Foundation of Aerospace SAST2021-012 and the National Natural Science Foundation of China (Grant Nos. 82103037, 82172860).
\end{acks}

\bibliographystyle{ACM-Reference-Format}
\bibliography{sample-base}

\end{document}